%% file: iclr2020.tex
\documentclass{article} 
\usepackage{iclr2020,times}
\usepackage{booktabs}
\usepackage{graphicx}

\input{math_commands.tex}

\usepackage{hyperref}
\usepackage{url}
\usepackage{bm}
\usepackage{subfig}
\usepackage{algorithm}
\usepackage[noend]{algpseudocode} 
\usepackage{array}
\newcolumntype{L}[1]{>{\raggedright\let\newline\\\arraybackslash\hspace{0pt}}m{#1}}
\newcolumntype{C}[1]{>{\centering\let\newline\\\arraybackslash\hspace{0pt}}m{#1}}
\newcolumntype{R}[1]{>{\raggedleft\let\newline\\\arraybackslash\hspace{0pt}}m{#1}}

\title{Trying again instead of trying longer: Prior learning for Automatic Curriculum Learning}

\author{
    \hspace{-0.07cm}R\'emy Portelas \\
    INRIA (FR)\\
    \texttt{remy.portelas@inria.fr}\\
  \And
    Katja Hofmann \\
    Microsoft Research (UK)\\
  \And
    Pierre-Yves Oudeyer \\
    INRIA (FR)\\
  }

\iclrfinalcopy 
\begin{document}

\maketitle
\begin{abstract}
A major challenge in the Deep RL (DRL) community is to train agents able to generalize over unseen situations, which is often approached by training them on a diversity of tasks (or environments). A powerful method to foster diversity is to procedurally generate tasks by sampling their parameters from a multi-dimensional distribution, enabling in particular to propose a different task for each training episode. In practice, to get the high diversity of training tasks necessary for generalization, one has to use complex procedural generation systems. With such generators, it is hard to get prior knowledge on the subset of tasks that are actually learnable at all (many generated tasks may be unlearnable), what is their relative difficulty and what is the most efficient task distribution ordering for training. A typical solution in such cases is to rely on some form of Automated Curriculum Learning (ACL) to adapt the sampling distribution. One limit of current approaches is their need to explore the task space to detect progress niches over time, which leads to a loss of time. Additionally, we hypothesize that the induced noise in the training data may impair the performances of brittle DRL learners. We address this problem by proposing a two stage ACL approach where 1) a teacher algorithm first learns to train a DRL agent with a high-exploration curriculum, and then 2) distills learned priors from the first run to generate an "expert curriculum" to re-train the same agent \textit{from scratch}. Besides demonstrating 50\% improvements on average over the current state of the art, the objective of this work is to give a first example of a new research direction oriented towards refining ACL techniques over multiple learners, which we call \textit{Classroom Teaching}.

\end{abstract}

\section{Introduction}
\paragraph{Automatic CL.} The idea of organizing the learning sequence of a machine is an old concept that stems from multiple works in reinforcement learning \citep{selfridge,Schmid}, developmental robotics \citep{oudeyer2007intrinsic} and supervised learning \citep{elman, bengiocl}, from which the Deep RL community borrowed the term \textit{Curriculum Learning} (CL). Automatic CL refers to approaches able to autonomously adapt their task sampling distribution to their evolving learner with minimal expert knowledge. Several ACL approaches have recently been proposed \citep{goalgan, settersolver,OpenAI2019SolvingRC,portelas2019,curious,skewfit, metarl-carml, Finot2019}.

\paragraph{ACL and exploration.} One of the limits of ACL is that when applied to a large parameterized task space with few learnable subspaces, as when considering a rich procedural generation system, they loose a lot of time finding the "optimal parameters" at a given point in time (e.g. the niches of progress in Learning Progress-based approaches) through \textit{task exploration}. We also hypothesize that these additional tasks presented to the DRL learner have a cluttering effect on the gathered training data, which adds noise in its already brittle gradient-based optimization and leads to sub-optimal performances. 

\paragraph{Proposed approach.} Given this hypothesized drawback of task exploration, we propose to study whether ACL techniques could be improved by having a two stage approach consisting in 1) a preliminary run with ACL from which prior knowledge on the task space is extracted, and 2) a second independent run leveraging this prior knowledge to propose a better curriculum to the DRL agent. For its simplicity and versatility, we choose to develop such an approach with ALP-GMM \citep{portelas2019}, a recent ACL algorithm for continuous task spaces.

\paragraph{Related work.} Within DRL, \textit{Policy Distillation} \citep{pol-dil-review} consists in leveraging a previously trained policy, the "teacher", and use it to perform \textit{behavior cloning} by training a "student" policy to jointly maximize its reward on one or several tasks while minimizing the distance between its action distribution compared to the teacher's. This allows to speed up the learning of bigger architectures and/or to leverage task-experts to train a single learner on a set of tasks. From this point of view, this work can be seen as a complementary approach interested in how to perform \textit{Curriculum Distillation} when considering a continuous space of tasks.

Similar ideas where developed for supervised learning by \cite{hacohen19a-score-pace}. In their work, authors propose an approach to infer a curriculum from past training for an image classification task: they use a first network trained without curriculum and use its predictive confidence for each image as a difficulty measure that a subsequently trained network uses for curriculum generation.
The idea that a knowledge distillation procedure can be beneficial even when the teacher and student policies have identical architectures has also been studied in supervised learning \citep{ban, banlike}. In this work, we propose to extend these concepts to DRL scenarios.

\section{Methods}
\label{methods}

\begin{figure*}[htb!]
\centering
\hspace{-0.08cm}\subfloat{\includegraphics[height=3.4cm]{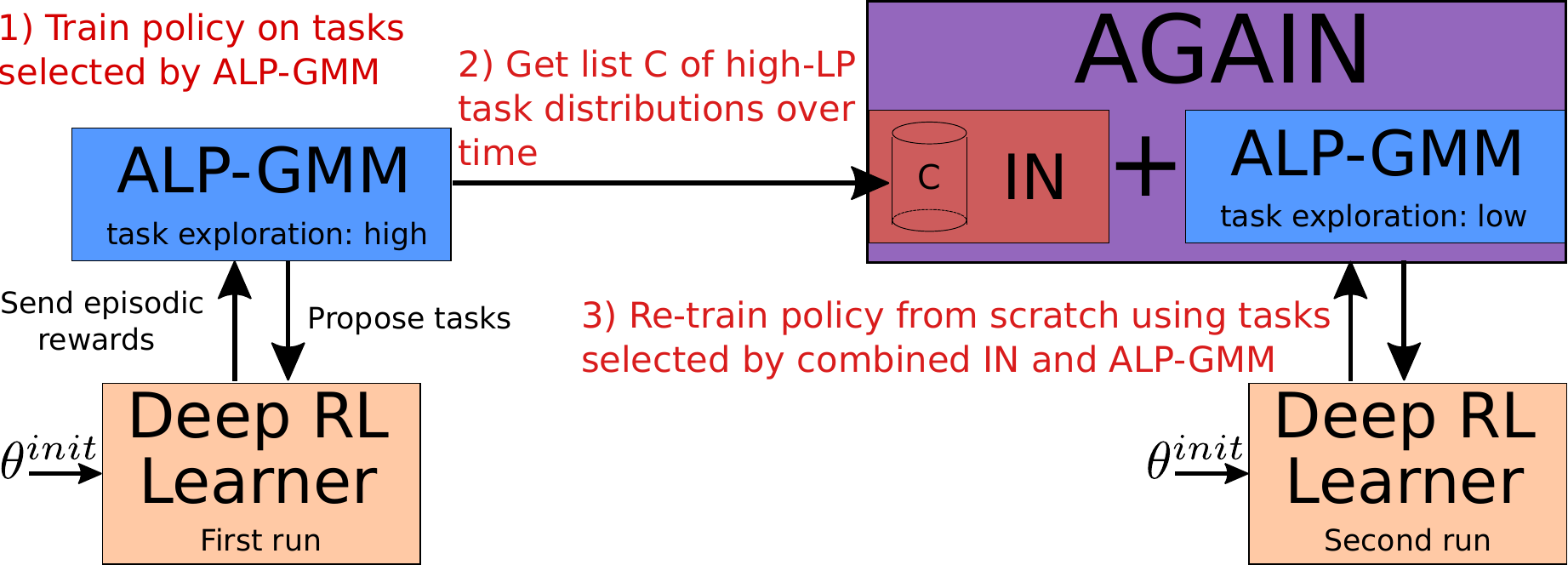}}
\hspace{0.3cm}\subfloat{\includegraphics[height=3.5cm]{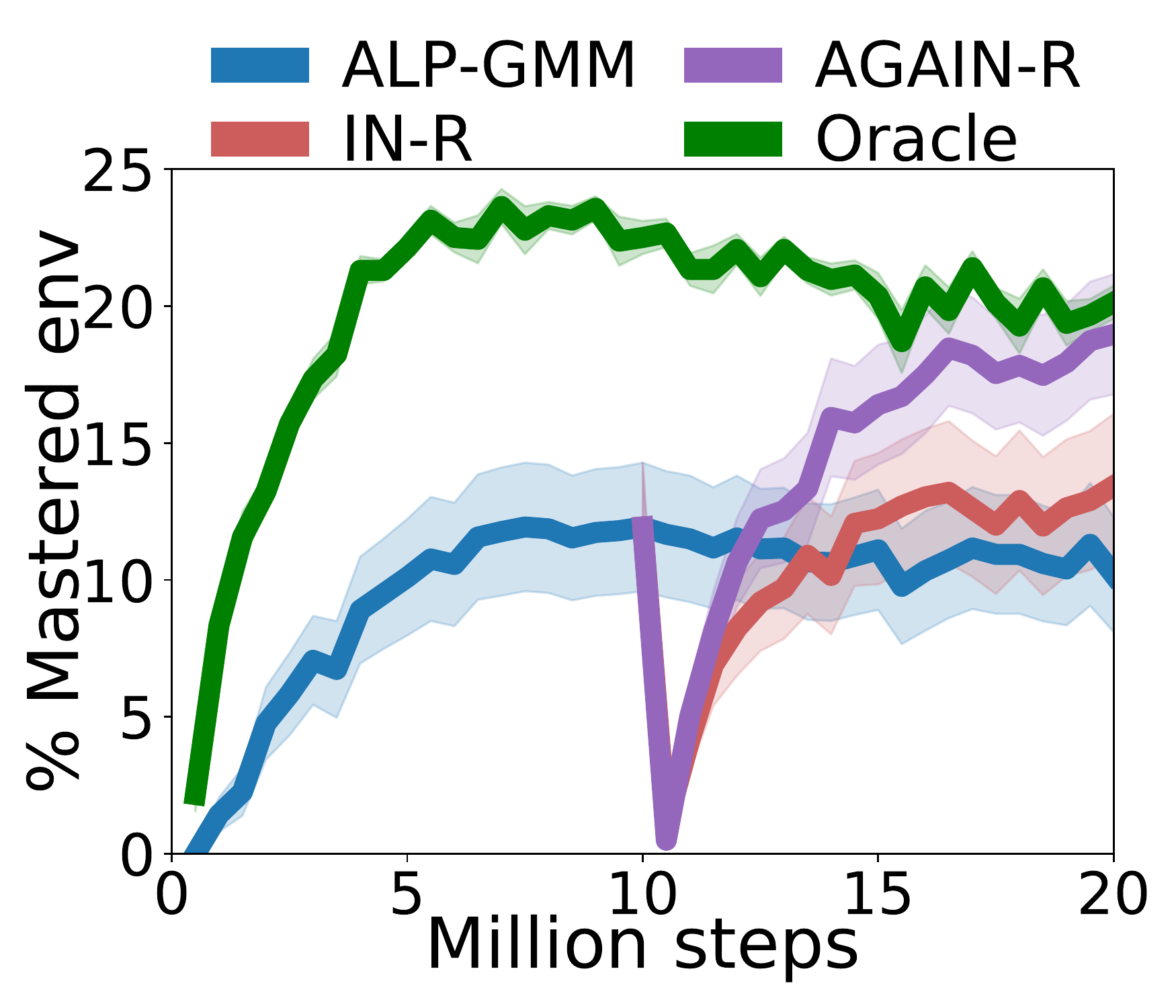}}
\caption{\footnotesize{\textbf{Left:} Schematic pipeline of Alp-Gmm And Inferred progress Niches (AGAIN), our proposed approach, which first leverages a preliminary run with a high-exploration ALP-GMM curriculum generator to infer an expert curriculum combined with a low-exploration ALP-GMM in a second run with the initial Deep RL learner, re-trained from scratch.} \textbf{Right:} Given identical training time, our combined approach outperforms regular ALP-GMM and even matches Oracle curriculum in a complex parametric BipedalWalker environment.}
\label{again-fig}
\end{figure*}


\paragraph{ALP-GMM} ALP-GMM \citep{portelas2019} is a Learning Progress (LP) based ACL technique for continuous task spaces that does not assume prior knowledge on the task space. It is inspired by previous works in developmental robotics \citep{riac, moulinfriergmm}. ALP-GMM frames the task sampling problem into an EXP-4 non-stationary Multi-Armed bandit setup \citep{auer2002nonstochastic} in which arms are Gaussians spanning over the space of tasks' parameters whose utility is defined with a local LP measure. The essence of ALP-GMM is to periodically fit a Gaussian Mixture Model (GMM) on recently sampled tasks' parameters \textit{concatenated with their respective LP}. Then, the Gaussian from which to sample a new task is chosen proportionally to its mean LP dimension. Task exploration happens initially through a bootstrapping period of random task sampling and during training by occasional random task sampling with probability $\rho_{rnd}$.
\paragraph{Inferred progress Niches (IN)} Using ALP-GMM is convenient for our target experiments as deriving an expert curriculum from an initial run is straightforward: one simply needs to gather the sequence of GMMs $\mathcal{C}_{raw}$ that were periodically fitted along training:
\begin{equation}
\label{eq:craw}
    C_{raw} = \{p(1), ..., p(T)\} \mid p(t)= \sum_{i=1}^{K_t} LP_{kt}\mathcal{N}(\bm{\mu_{ti}},\bm{\Sigma_{ti}}),
\end{equation}
with $T$ the total number of GMMs in the list, $K_t$ their respective number of components and $LP_{kt}$ the Learning Progress of each Gaussian. By keeping only Gaussians with $LP_{kt}$ above a predefined threshold $\delta_{LP}$, we can get a curated list $\mathcal{C}$. We name the resulting ACL approach Infered progress Niches (IN) and propose $3$ variants to select which GMM from $\mathcal{C}$ is used to sample tasks over time during the second run:
\begin{itemize}
    \item \textbf{Pool-based (IN-P)} A rather crude approach is to disregard the ordering of $\mathcal{C}$ and consider the entire trajectory of GMMs as one single pool $P$ of Gaussians, ie. one big mixture having $K_p=\sum_{t=1}^{T} K_t$ components.
    \item \textbf{Time-based (IN-T)} In this version $\mathcal{C}$ is stepped in periodically at the same rate than the preliminary ALP-GMM run (ie. a new GMM every $250$ episodes in our experiments).
    \item \textbf{Reward-based (IN-R)} Another option is to iterate over $\mathcal{C}$ only once the mean episodic reward over tasks recently sampled from the current GMM matches or surpasses the mean episodic reward recorded during the initial run (on the same GMM).
\end{itemize}
Regardless the selection process, given a GMM, a new task is selected by sampling a tasks' parameter on a Gaussian selected proportionally to its $LP_{kt}$ value.

\paragraph{Mixing both in AGAIN.} Simply using one of the $3$ proposed IN algorithms directly for the second run lacks adaptive mechanisms towards the characteristics of the second agent, whose initial parameters and data training stream are different from the first, which could lead to failure cases where the expert curriculum and the second learner are no longer "in phase". Additionally, if the initial run failed to discover progress niches, IN is bound to fail. As such we propose to combine IN with an ALP-GMM teacher (with low task-exploration) in the second run. The resulting Alp-Gmm And Inferred progress Niches approach, AGAIN for short, samples tasks from a GMM that is composed of the current mixture of both ALP-GMM and IN. See figure \ref{again-fig} for a schematic pipeline and appendix \ref{app-again} for details.

\section{Experiments and Results}
\newcounter{example}[section] 
\refstepcounter{example}

\paragraph{Evaluation procedure.} We propose to test our considered variants and baselines on a parametric version of BipedalWalker proposed by \cite{portelas2019},  which generates walking tracks paved with stumps whose height and spacing are defined by a $2$-D parameter vector used for the procedural generation of tasks. This continuous task space has boundaries set in such a way that a substantial part of the space consists in unfeasible tracks for the default walker. As in their work, we also test our approaches with a modified short-legged walker, which constitutes an even more challenging scenario (as the task space is unchanged). All ACL variants are tested when paired with a Soft-Actor Critic \citep{sac} policy. Performance is measured by tracking the percentage of mastered tasks from a fixed test set. See appendix \ref{an:details} for details.

\paragraph{Is re-training from scratch beneficial?} The end performances of all tested conditions are summarized in table \ref{results-table}. Interestingly, for all tested variants, retraining the DRL agent from scratch in the second run gave superior end performances than fine-tuning using the weights of the first run \textit{in all tested variants}. This showcase the brittleness of gradient-based training and the difficulty of transfer learning. Despite this, even fine-tuned variants reached superior end-performances than classical ALP-GMM, meaning that the change in curriculum strategy in itself is already beneficial. 

\paragraph{Is it useful to re-use ALP-GMM in the second run?} In the default walker experiments, AGAIN-R, T and P conditions mixing ALP-GMM and IN in the second run reached lower mean performances than their respective IN variants. However, the exact opposite is observed for IN-R and IN-T variants in the short walker experiments. This can be explained by the difficulty of short walker experiments for ACL approaches, leading to $16/30$ preliminary 10M steps long ALP-GMM runs to have a mean end-performance of $0$, compared to $0/30$ in the default walker experiments. All these run failures led to many GMMs lists $\mathcal{C}$ used in IN to be of very low-quality, which illustrates the advantage of AGAIN that is able to emancipate from IN using ALP-GMM.



\paragraph{Highest-performing variants.} Consistently with the precedent analysis, mixing ALP-GMM with IN in the second run is not essential in default walker experiments, as the best performing ACL approach is IN-P. This most likely suggests that the improved adaptability of the curriculum when using AGAIN is outbalanced by the added noise (due to the low task-exploration). However in the more complex short walker experiments, mixing ALP-GMM with IN is essential, especially for AGAIN-R, which substantially outperforms ALP-GMM and other AGAIN and IN variants, reaching a mean end performance of $19.0$. The difference in end-performance between AGAIN-R and Oracle, our hand-made expert using privileged information who obtained $20.1$, is not statistically significant ($p=0.6$).

\vspace{0.5cm}
\begin{minipage}[t]{0.6\textwidth}
\begin{tabular}{@{}lll@{}}
\toprule
Condition            & Short walker & Default walker \\ \midrule
AGAIN-R               & $19.0\pm12.0^*$               & $41.6\pm6.3^*$           \\
AGAIN-R(fine-tune)    & $11.4\pm12.9$                 & $39.9\pm4.6$           \\
IN-R                  & $13.4\pm14.4$                 & $43.5\pm9.6^*$           \\
IN-R(fine-tune)       & $11.2\pm12.3$                 & $40.8\pm5.6$           \\
AGAIN-T               & $15.1\pm11.9$                 & $40.6\pm11.5$           \\
AGAIN-T(fine-tune)    & $11.4\pm11.8$                 & $40.6\pm3.8^*$           \\
IN-T                  & $13.5\pm13.3$                 & $43.5\pm6.1^*$           \\
IN-T(fine-tune)       & $10.7\pm12.3$                 & $40.3\pm7.6$           \\
AGAIN-P               & $13.6\pm12.5$                 & $41.9\pm5.1^*$           \\
AGAIN-P(fine-tune)    & $11.1\pm12.0$                 & $41.5\pm3.9^*$           \\
IN-P                  & $14.5\pm12.6$                 & $\mathbf{44.3}\pm3.5^*$  \\
IN-P(fine-tune)       & $12.2\pm12.5$                 & $41.1\pm3.8^*$           \\
ALP-GMM               & $10.2\pm11.5$                 & $38.6\pm3.5$           \\
Oracle                & $\mathbf{20.1}\pm3.4^*$         & $27.2\pm15.2^-$           \\
Random                & $2.5\pm5.9^-$                   & $20.9\pm11.0^-$           \\ \bottomrule
\label{results-table}
\end{tabular}
\end{minipage}
\begin{minipage}[t]{0.4\textwidth}
  \vspace{-2.55cm}
  Table 1: \footnotesize{\textbf{Experiments on Stump Tracks with short and default bipedal walkers.} The average performance with standard deviation after 10 Millions steps (IN and AGAIN variants) or 20 Million steps (others) is reported (30 seeds per condition). For IN and AGAIN we also test variants that do not retrain the weights of the policy used in the second run \textit{from scratch} but rather \textit{fine-tune} them from the preliminary run.$\mathbf{^{*/-}}$ Indicates whether performance difference with ALP-GMM is statistically significant ie. $p<0.05$ in a post-training Welch's student t-test ($\mathbf{^{*}}$ for performance advantage w.r.t ALP-GMM and $\mathbf{^{-}}$ for performance disadvantage).}
\end{minipage}

\section{Conclusion and Discussion}

In this work we presented Alp-Gmm And Inferred progress Niches, a simple yet effective approach to learn prior knowledge over a space of tasks to design a curriculum tailored to a DRL agent. Instead of following the same exploratory ACL approach over the entire training, AGAIN performs a first preliminary run with ALP-GMM, derives a list of progress niches from it, and uses this list to build an expert curriculum that is combined with a low task-exploration ALP-GMM teacher for a second run of the same DRL agent, trained from scratch.

\paragraph{Beyond tabula rasa?} In this work we showed that a non-tabula rasa curriculum generator that leveraged prior knowledge over the task space (from a preliminary run) outperformed the regular approach that learned to generate an entire curriculum from scratch. However, we also demonstrated that, from the point of view of the DRL learner, it is actually \textit{better} to restart tabula rasa (with a non-tabula rasa curriculum generator), which is a very interesting perspective and opens several lines for future work.

\paragraph{Classroom Teaching} Beyond proposing a two-stage ACL technique for a single DRL agent, the experimental setup of this work could be seen as a particular case of a broader problem we propose to name \textit{Classroom Teaching} (CT). CT defines a family of problems in which a \textit{meta-ACL} algorithm is tasked to either sequentially or simultaneously generate multiple curricula tailored for each of the learning students, all having potentially varying abilities. CT differs from the problems studied in population-based developmental robotics \citep{imgep} and evolutionary algorithms \citep{poet} as in CT the number and characteristics of learners are predefined, and the objective is to foster maximal learning progress over all learners rather than iteratively constructing high-performing policies. Studying CT scenarios brings DRL closer to human education research problems and might stimulate the design of methods that alleviate the expensive use of expert knowledge in current state of the art assisted education \citep{zpdes, Koedinger13}.



\bibliography{iclr2020}
\bibliographystyle{iclr2020}

\clearpage
\appendix
\section{ALP-GMM}
\label{app-alp-gmm}
\begin{figure*}
\centering
\includegraphics[width=\textwidth]{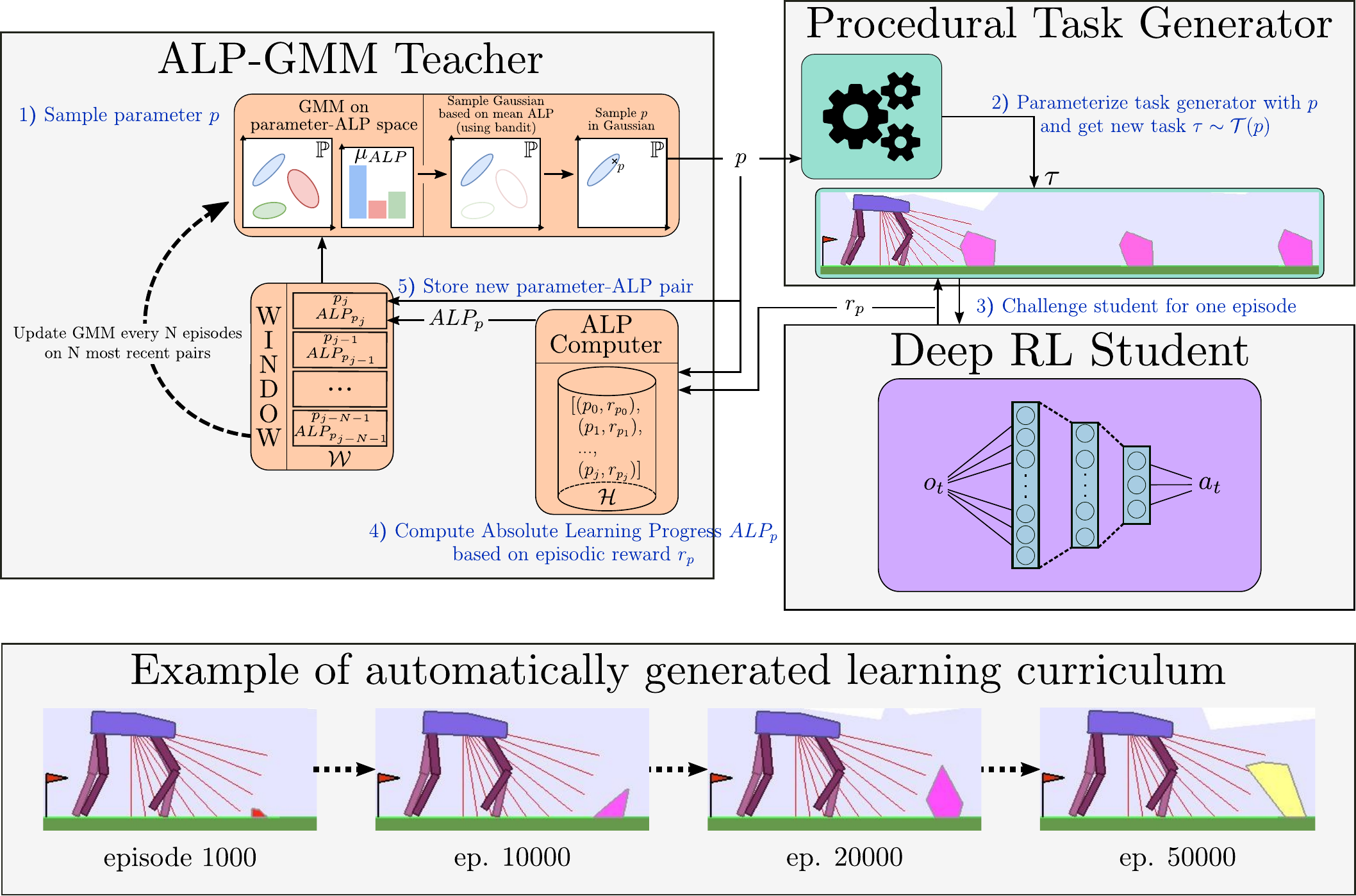}
\caption{\footnotesize{Schematic view of an ALP-GMM teacher's workflow from \cite{portelas2019}}}
\label{ALP-GMM-pipeline}
\end{figure*}

ALP-GMM relies on an empirical per-task computation of Absolute Learning Progress (ALP), allowing to fit a GMM on a concatenated space composed of tasks' parameters and respective ALP. Given a task $\tau_{new} \in \mathcal{T}$ whose parameter is $p_{new} \in \mathcal{P}$ and on which the policy collected the episodic reward $r_{new} \in \mathbb{R}$, Its ALP is computed using the closest previous tasks $\tau_{old}$ (Euclidean distance) with associated episodic reward $r_{old}$:
\begin{equation}
    \label{eq:2}
    alp_{new} = |r_{new} - r_{old}|
\end{equation}
All previously encountered task's parameters and their associated ALP, parameter-ALP for short, recorded in a history database $\mathcal{H}$, are used for this computation. Contrastingly, the fitting of the GMM is performed every $N$ episodes on a window $\mathcal{W}$ containing the $N$ most recent parameter-ALP. The resulting mean ALP dimension of each Gaussian of the GMM is used for proportional sampling. To adapt the number of components of the GMM online, a batch of GMMs having from 2 to $k_{max}$ components is fitted on $\mathcal{W}$, and the best one, according to Akaike's Information Criterion \citep{aic}, is kept as the new GMM. In our experiments we use the same hyperparameters as in \cite{portelas2019} ($N=250$, $k_{max}=10$), except for the percentage of random task sampling $\rho_{rnd}$ which we set to $10\%$ (we found it to perform better than $20\%$) when running ALP-GMM alone or $2\%$ when combined with IN in the second phase of AGAIN. See algorithm \ref{algo:ALP-GMM} for pseudo-code and figure \ref{ALP-GMM-pipeline} for a schematic pipeline. Note that in this paper we refer to ALP as LP for simplicity (ie. $LP_{kt}$ in $\mathcal{C}$ from eq. \ref{eq:craw} is equivalent to the mean ALP of Gaussians in ALP-GMM).

\begin{algorithm}[H]
	\caption{~ Absolute Learning Progress Gaussian Mixture Model (ALP-GMM)}
	\label{algo:ALP-GMM}
	\begin{algorithmic}[1]
	
	\Require Student policy $\pi_\theta$, parametric procedural environment generator $E$, bounded parameter space $\mathcal{P}$, probability of random sampling $\rho_{rnd}$, fitting rate $N$, max number of Gaussians $k_{max}$
	\vspace{0.2cm}
	\State Initialize $\pi_\theta$
	\State Initialize parameter-ALP First-in-First-Out window $\mathcal{W}$, set max size to $N$
	\State Initialize parameter-reward history database $\mathcal{H}$
	\Loop~$N$ times \Comment Bootstrap phase
	    \State Sample random $p \in \mathcal{P}$, send $E(\tau \sim \mathcal{T}(p))$ to $\pi_\theta$, observe episodic reward $r_p$
	    \State Compute ALP of $p$ based on $r_p$ and $\mathcal{H}$ (see equation \ref{eq:2})
	    \State Store $(p,r_p)$ pair in $\mathcal{H}$, store $(p,ALP_p)$ pair in $\mathcal{W}$
	\EndLoop
	\Loop \Comment Stop after $K$ inner loops
	\State Fit a set of GMM having 2 to $k_{max}$ kernels on $\mathcal{W}$
	\State Select the GMM with best Akaike Information Criterion
	\Loop~$N$ times
	    \State $\rho_{rnd} \%$ of the time, sample a random parameter $p \in \mathcal{P}$
	    \State Else, sample $p$ from a Gaussian chosen proportionally to its mean ALP value 
		\State Send $E(\tau \sim \mathcal{T}(p))$ to student $\pi_\theta$ and observe episodic reward $r_p$
	    \State Compute ALP of $p$ based on $r_p$ and $\mathcal{H}$
	    \State Store $(p,r_p)$ pair in $\mathcal{H}$, store $(p,ALP_p)$ pair in $\mathcal{W}$
	\EndLoop
	\EndLoop
	\State \textbf{Return} $\pi_\theta$
	
	\end{algorithmic}
\end{algorithm}

\clearpage
\section{AGAIN}
\label{app-again}

\paragraph{IN variants.} In order to filter the list $\mathcal{C}_{raw}$ (see eq. \ref{eq:craw}) of GMMs collected after a preliminary run of ALP-GMM into $\mathcal{C}$ and use it as an expert curriculum, we remove any Gaussian with a $LP_{kt}$ below $\delta_{LP}=0.1$ (the LP dimension is normalized between $0$ and $1$, which requires to choose an approximate potential reward range, set to $[-150,350]$ for all experiments). When all Gaussians of a GMM are discarded, the GMM is removed from $\mathcal{C}$. In practice, it allows to remove non-informative GMMs corresponding to the initial exploration phase of ALP-GMM, when the learner has not made any progress (hence no LP detected by the curriculum generator). $\mathcal{C}$ is then iterated over to generate a curricula with either of the Time-based (see algo \ref{int-algo}), Pool-based (see algo \ref{inp-algo}) or Reward-based (see algo \ref{inr-algo}) IN. The IN-P approach does not require additional hyperparameters. The IN-T requires an update rate $N$ to iterate over $\mathcal{C}$, which we set to $250$ (same as the fitting rate of ALP-GMM). The IN-R approach requires to extract additional data from the first run, in the form of a list $\mathcal{R}_{raw}$:
\begin{equation}
\label{eq:rraw}
    R_{raw} = \{\mu_r^1, ...,\mu_r^t, \mu_r^T\} ~~s.t~~~ |\mathcal{R}_{raw}| = |\mathcal{C}_{raw}|,
\end{equation}
with T the total number of GMMs in the first run (same as in $\mathcal{C}_{raw}$), and $\mu_r^t$ the mean episodic reward obtained by the first DRL agent during the last $50$ tasks sampled from the $t^{th}$ GMM. $\mathcal{R}$ is simply obtained by removing any $\mu_r^t$ that corresponds to a GMM discarded while extracting $\mathcal{C}$ from $\mathcal{C}_{raw}$. The remaining rewards are then used as thresholds in IN-R to decide when to switch to the next GMM in $\mathcal{C}$.

\paragraph{AGAIN} In AGAIN (see algo. \ref{again-algo}), the idea is to use both IN (R,T or P) and ALP-GMM (without the random bootstrapping period) for curriculum generation. We combine the changing GMM of IN and ALP-GMM over time, simply by building a GMM $G$ containing Gaussians from the current GMM of IN and ALP-GMM. By selecting the Gaussian in $G$ from which to sample a new task using their respective LP, This approach allows to adaptively modulate the task sampling between both, shifting the sampling towards IN when ALP-GMM does not detect high-LP subspaces and towards ALP-GMM when the current GMM of IN has low-LP Gaussians. Additionally, to have minimal task-exploration, which benefits ALP-GMM (allowing it to detect new progress niches), we sample random tasks with probability $\rho_{low}=2\%$ (compared with $\rho_{high}=10\%$ used for the preliminary ALP-GMM run). 

\begin{algorithm}[H]
	\caption{~ Inferred progress Niches - Time-based (IN-T)}
	\label{int-algo}
	\begin{algorithmic}[1]
	
	\Require Student policy $\mathcal{\pi_\theta}$, task-encoding parameter space $\mathcal{P}$, LP threshold $\delta_{LP}$, update rate $N$, experimental budget $K$, random sampling ratio $\rho_{high}$, parametric procedural environment generator $E$
	\vspace{0.2cm}
	\State Init $\mathcal{\pi_\theta}$, train it for $K/2$ env. steps with ALP-GMM($\rho_{high}, \mathcal{P}$) and collect $\mathcal{C}_{raw}$ \Comment First run
	\State Get $\mathcal{C}$ from $\mathcal{C}_{raw}$ by removing any Gaussian with $LP_{kt} < \delta_{LP}$.
	\State Re-initialize $\mathcal{\pi_\theta}$ \Comment Second run
	\State Initialize expert curriculum index $i_{c}$ to $0$
	\Loop~  \Comment Stop after $K/2$ environment steps
	    \State Set $i_{c}$ to $min(i_{c}+1, len(\mathcal{C}))$
	    \State Set current GMM $G_{IN}$ to $i_{c}^{th}$ GMM in $\mathcal{C}$
	    \Loop~$N$ times
	    \State Sample $p$ from a Gaussian in $G_{IN}$ chosen proportionally to its $LP_{kt}$
		\State Send $E(\tau \sim \mathcal{T}(p))$ to student $\mathcal{\pi_\theta}$
	\EndLoop
	\EndLoop
	\State \textbf{Return} $\mathcal{\pi_\theta}$
	\end{algorithmic}
\end{algorithm}

\begin{algorithm}[H]
	\caption{~ Inferred progress Niches - Pool-based (IN-P)}
	\label{inp-algo}
	\begin{algorithmic}[1]
	
	\Require Student policy $\mathcal{\pi_\theta}$, task-encoding parameter space $\mathcal{P}$, LP threshold $\delta_{LP}$, experimental budget $K$, random sampling ratio $\rho_{high}$, parametric procedural environment generator $E$
	\vspace{0.2cm}
	\State Init $\mathcal{\pi_\theta}$, train it for $K/2$ env. steps with ALP-GMM($\rho_{high}, \mathcal{P}$) and collect $\mathcal{C}_{raw}$ \Comment First run
	\State Get $\mathcal{C}$ from $\mathcal{C}_{raw}$ by removing any Gaussian with $LP_{kt} < \delta_{LP}$.
	\State Re-initialize $\mathcal{\pi_\theta}$ \Comment Second run
	\State Initialize pool GMM $G_{IN}$, containing all Gaussians from $\mathcal{C}$
	\Loop~  \Comment Stop after $K/2$ environment steps
	    \State Sample $p$ from a Gaussian in $G_{IN}$ chosen proportionally to its $LP_{kt}$
		\State Send $E(\tau \sim \mathcal{T}(p))$ to student $\mathcal{\pi_\theta}$
	\EndLoop
	\State \textbf{Return} $\mathcal{\pi_\theta}$
	\end{algorithmic}
\end{algorithm}

\begin{algorithm}[H]
	\caption{~ Inferred progress Niches - Reward-based (IN-R)}
	\label{inr-algo}
	\begin{algorithmic}[1]
	
	\Require Student policy $\mathcal{\pi_\theta}$, task-encoding parameter space $\mathcal{P}$, LP threshold $\delta_{LP}$, memory size $N$, experimental budget $K$, random sampling ratio $\rho_{high}$, parametric procedural environment generator $E$
	\vspace{0.2cm}
	\State Init $\mathcal{\pi_\theta}$, train it for $K/2$ env. steps with ALP-GMM($\rho_{high}, \mathcal{P}$) and collect $\mathcal{C}_{raw}$ \Comment First run
	\State Get $\mathcal{C}$ from $\mathcal{C}_{raw}$ by removing any Gaussian with $LP_{kt} < \delta_{LP}$.
	\State Additionally, collect list of inferred reward thresholds $\mathcal{R}_{raw}$, and get $\mathcal{R}$ \Comment See eq. \ref{eq:rraw}
	\State Re-initialize $\mathcal{\pi_\theta}$ \Comment Second run
	\State Initialize reward First-in-First-Out window $\mathcal{W}$, set max size to $N$
	\State Initialize expert curriculum index $i_{c}$ to $0$
	\Loop~  \Comment Stop after $K/2$ environment steps
	    \State If $\mathcal{W}$ is full, compute mean reward $\mu_w$ from $\mathcal{W}$
	    \State ~~~~If $\mu_w$ superior to $i_{c}^{th}$ reward threshold in $\mathcal{R}$, set $i_{c}$ to $min(i_{c}+1, len(\mathcal{C}))$
	    \State Set current GMM $G_{IN}$ to $i_{c}^{th}$ GMM in $\mathcal{C}$
	    \State Sample $p$ from a Gaussian in $G_{IN}$ chosen proportionally to its $LP_{kt}$
		\State Send $E(\tau \sim \mathcal{T}(p))$ to student $\mathcal{\pi_\theta}$ and add episodic reward $r_p$ to $\mathcal{W}$
	\EndLoop
	\State \textbf{Return} $\mathcal{\pi_\theta}$
	\end{algorithmic}
\end{algorithm}

\begin{algorithm}[H]
	\caption{~ Alp-Gmm And Inferred progress Niches (AGAIN)}
	\label{again-algo}
	\begin{algorithmic}[1]
	
	\Require Student policy $\mathcal{\pi_\theta}$, task-encoding parameter space $\mathcal{P}$, random sampling ratio $\rho_{low}$ and $\rho_{high}$ , LP threshold $\delta_{LP}$, experimental budget $K$, parametric procedural environment generator $E$
	\vspace{0.2cm}
	\State Init $\mathcal{\pi_\theta}$, train it for $K/2$ env. steps with ALP-GMM($\rho_{high}, \mathcal{P}$) and collect $\mathcal{C}_{raw}$ \Comment First run
	\State Get $\mathcal{C}$ from $\mathcal{C}_{raw}$ by removing any Gaussian with $LP_{kt} < \delta_{LP}$
	\State re-initialize $\mathcal{\pi_\theta}$ \Comment Second run
	\State Setup new ALP-GMM($\rho=0, \mathcal{P}$) \Comment See algo. \ref{algo:ALP-GMM}
	\State Setup either IN-T, IN-P or IN-R \Comment See algo. \ref{int-algo}, \ref{inp-algo} and \ref{inr-algo}
	\Loop  \Comment Stop after $K/2$ environment steps
	    \State Get composite GMM $G$ from the current GMM of both ALP-GMM and IN
	    \State $\rho_{low} \%$ of the time, sample a random parameter $p \in \mathcal{P}$
	    \State Else, sample $p$ from a Gaussian chosen proportionally to its $LP$ 
		\State Send $E(\tau \sim \mathcal{T}(p))$ to student $\mathcal{\pi_\theta}$ and observe episodic reward $r_p$
	    \State Send $(p,r_p)$ pair to both ALP-GMM and IN
	\EndLoop
	\State \textbf{Return} $\mathcal{\pi_\theta}$
	
	\end{algorithmic}
\end{algorithm}

\clearpage
\section{Experimental details}
\label{an:details}

\paragraph{Soft Actor-Critic} In our experiments, we use an implementation of Soft Actor-Critic provided by OpenAI\footnote{https://github.com/openai/spinningup}. We use a $2$ layered ($400$,$300$) network for V, Q1, Q2 and the policy. Gradient steps are performed each $10$ environment steps, with a learning rate of $0.001$ and a batch size of $1000$. The entropy coefficient is set to $0.005$. 

\begin{figure*}[htb!]
\centering
\includegraphics[width=0.9\textwidth]{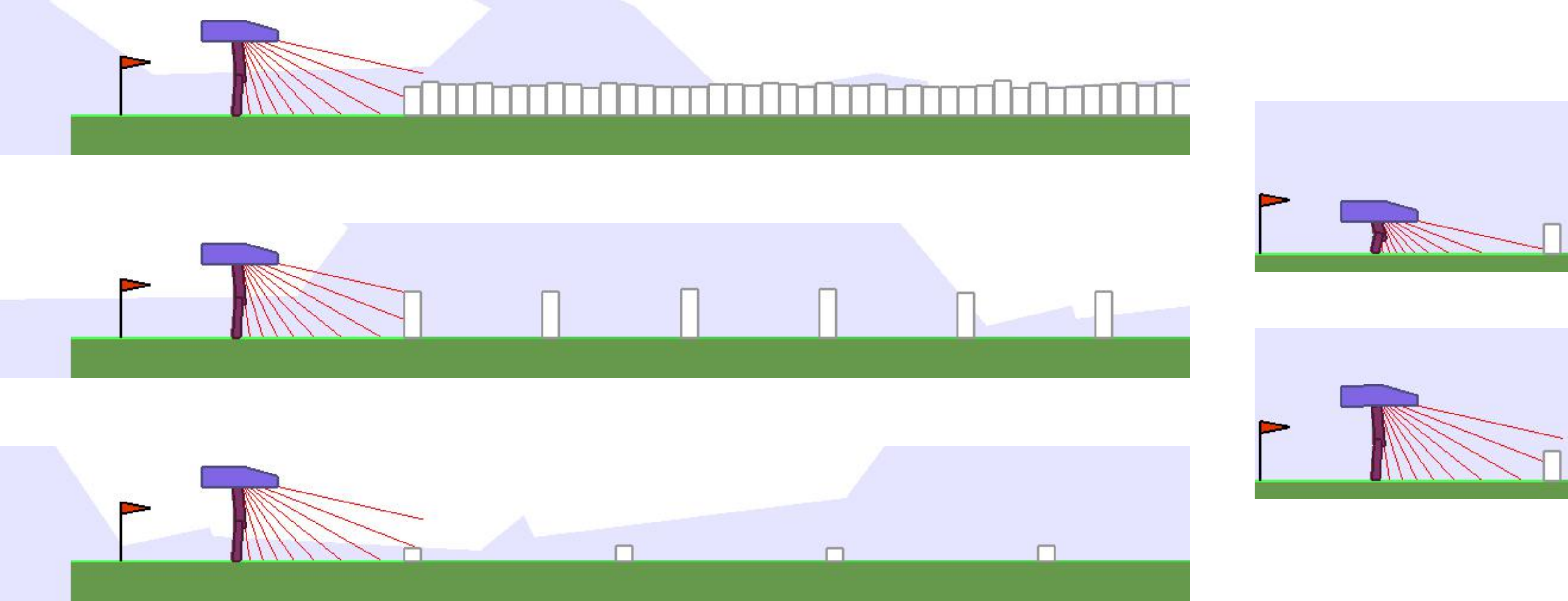}
\caption{\footnotesize{\textbf{Parameterized BipedalWalker environment.} \textbf{Left:} Examples of generated tracks. \textbf{Right:} The two walker morphologies tested on the environment.} One parameter tuple ($\mu_h, \delta_s$) actually encodes a \textit{distribution} of tasks as the height of each stump along the track is drawn from $\mathcal{N}(\mu_h,0.1)$. }
\label{pbw-demo}
\end{figure*}

\paragraph{Parametric BipedalWalker} Our proposed ACL variants choose parameters of tasks that encode the procedural generation of walking tracks paved with stumps in the BipedalWalker environments. As in \cite{portelas2019}, we bound the height dimension to $[0,3]$ and the spacing dimension to $[0,6]$ (regardless the walker morphology). The agent is rewarded for keeping its head straight and going forward and is penalized for torque usage. The episode is terminated after 1) reaching the end of the track, 2) reaching a maximal number of $2000$ steps, or 3) head collision (for which the agent receives a strong penalty). See figure \ref{pbw-demo} for visualizations.

\paragraph{Baselines} The Random curriculum baseline samples tasks' parameters randomly over the parameter space. The Oracle condition is a hand-made curriculum that is very similar to IN-R, except that the list $\mathcal{C}$ is built using expert knowledge, and all reward thresholds $\mu_r^i$ in $\mathcal{R}$ are set to $230$, which is an episodic reward value often used in the literature as characterizing a default walker having a "reasonably efficient" walking gate \citep{poet}. Basically, Oracle starts proposing tasks from a Gaussian (with std of $0.05$) located at the simplest subspace of the task space (ie. low stump height and high stump spacing) and then gradually moves the Gaussian towards the hardest subspaces (high stump height and low stump spacing) by small increments ($50$ steps overall) happening whenever the mean episodic reward of the DRL agent over the last $50$ proposed tasks is superior to $230$. In our experiments, consistently with \citep{portelas2019}, which implements a similar approach, Oracle is prone to forgetting due to the strong shift in task subspace (which is why it is not the best performing condition for default walker experiments (see table \ref{results-table}).

\paragraph{Computational resources.} To perform our experiments, we ran each condition for either $10$ (IN and AGAIN variants) or $20$ (others) Millions environment steps ($30$ repeats) using one cpu and one GPU (the GPU is shared between $8$ runs), for approximately 30 hours of wall-clock time. It amounts to $16200$ CPU-hours and $2025$ GPU-hours. The preliminary ALP-GMM runs used in IN and AGAIN variants correspond to the first $10$ Million steps of the ALP-GMM condition (whose end-performance after $20$ Million steps is reported in table \ref{results-table}.

\clearpage



\section{Additional Visualizations}
\begin{figure*}[htb!]
\centering
\subfloat[with \textbf{Pool-based} IN]{\includegraphics[width=0.9\textwidth]{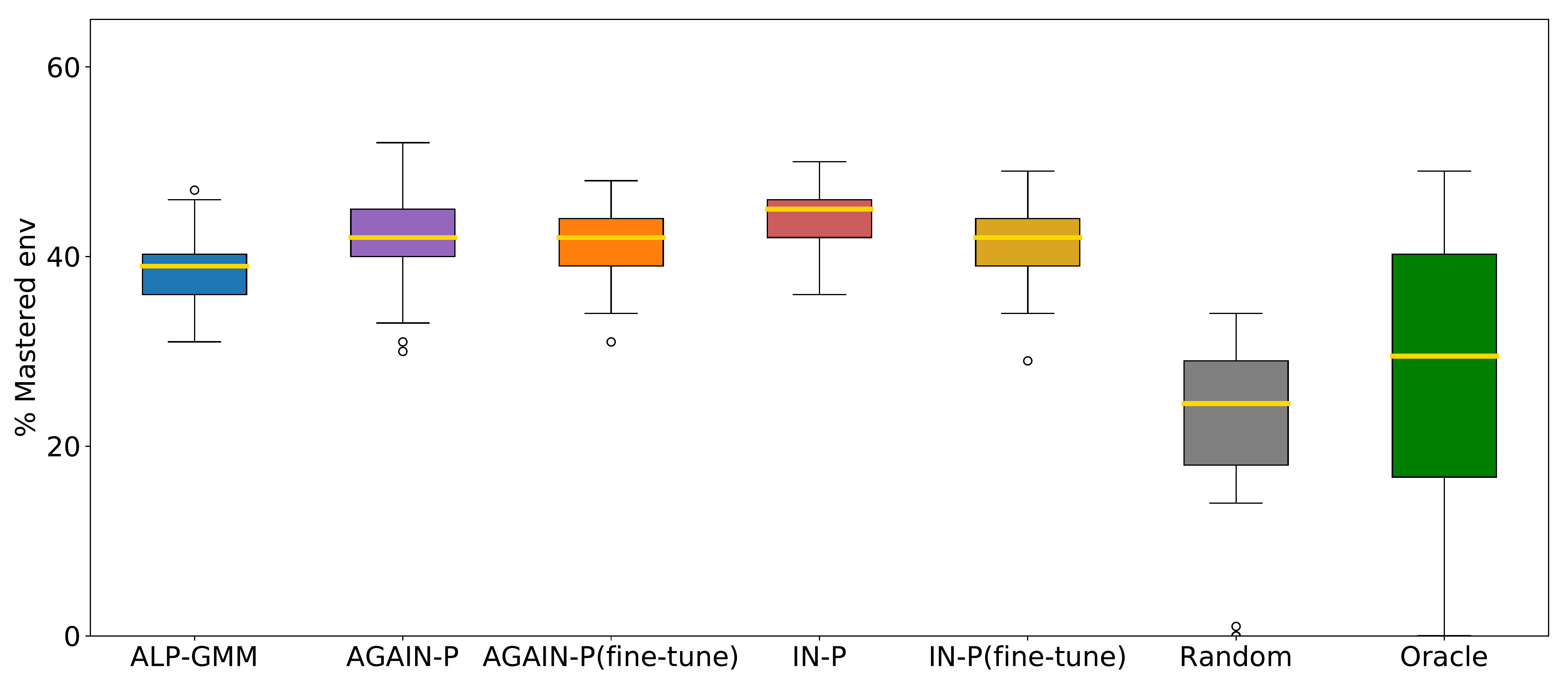}}

\subfloat[with \textbf{Time-based} IN]{\includegraphics[width=0.9\textwidth]{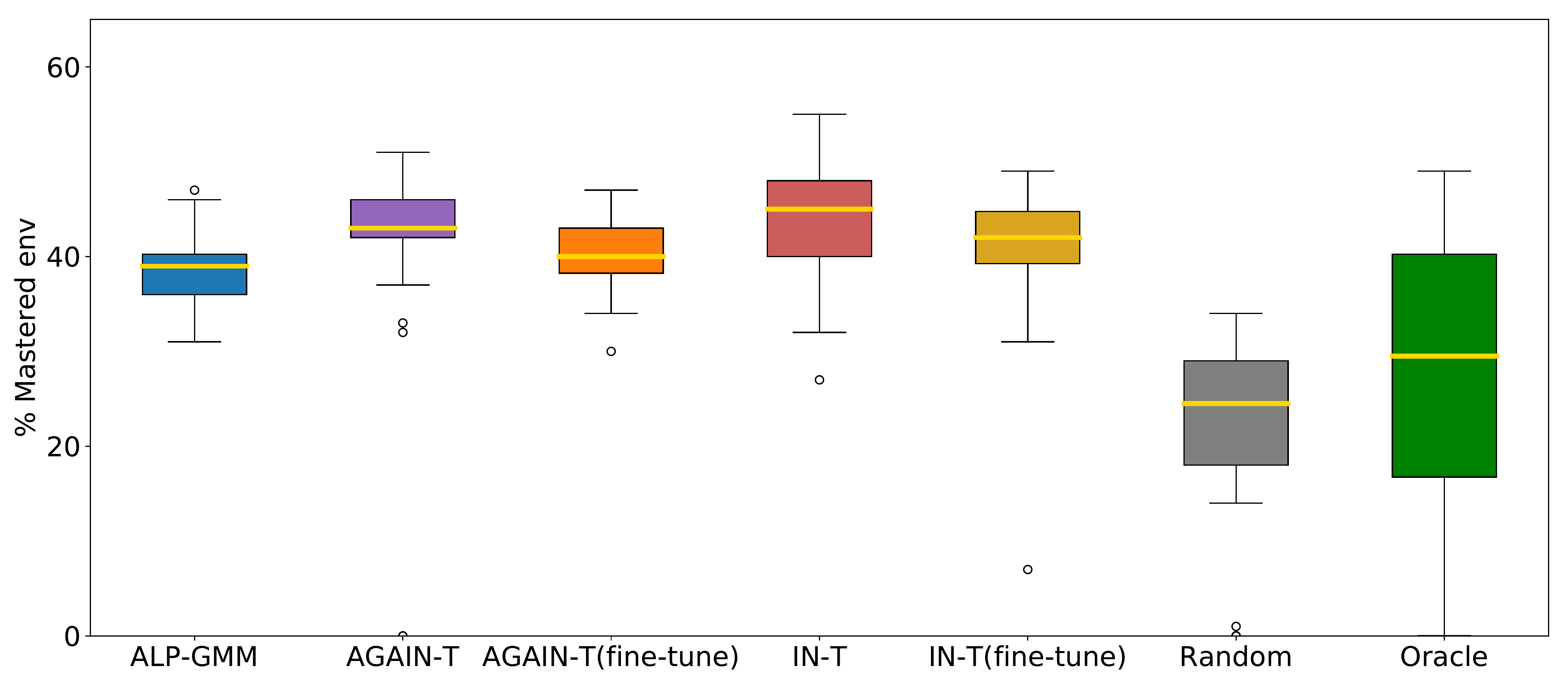}}

\subfloat[with \textbf{Reward-based} IN]{\includegraphics[width=0.9\textwidth]{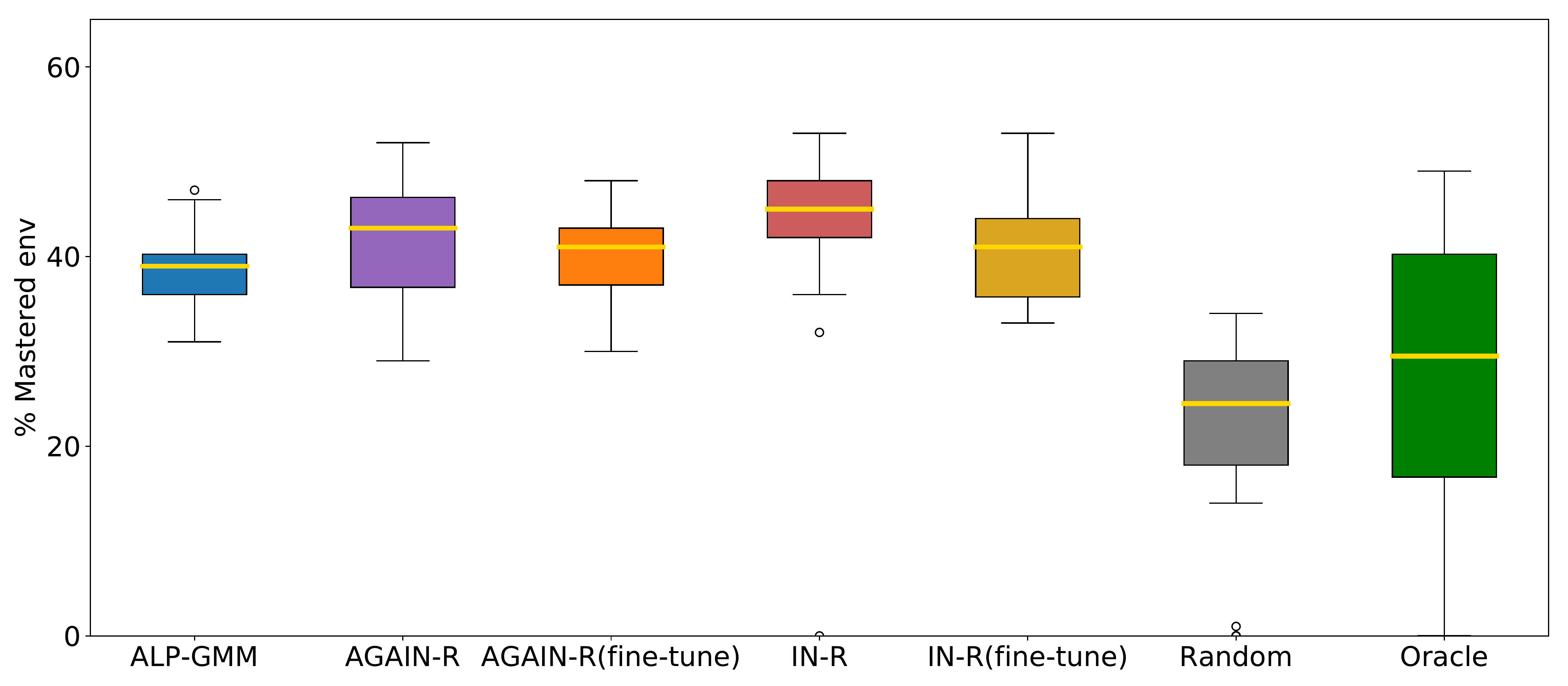}}
\caption{\footnotesize{\textbf{Box plots of the final performance of each condition with default bipedal walker after 20M environment steps.} Gold lines are medians, surrounded by a box showing the first and third quartile, which are then followed by whiskers extending to the last datapoint or $1.5$ times the inter-quartile range. Beyond the whiskers are outlier datapoints.} From top to bottom, each box plot presents results when using either Pool-based, Reward-based, or Time-based IN, respectively. }
\label{default-exps-boxplots}
\end{figure*}

\begin{figure*}[htb!]
\centering
\subfloat[with \textbf{Pool-based} IN]{\includegraphics[width=0.9\textwidth]{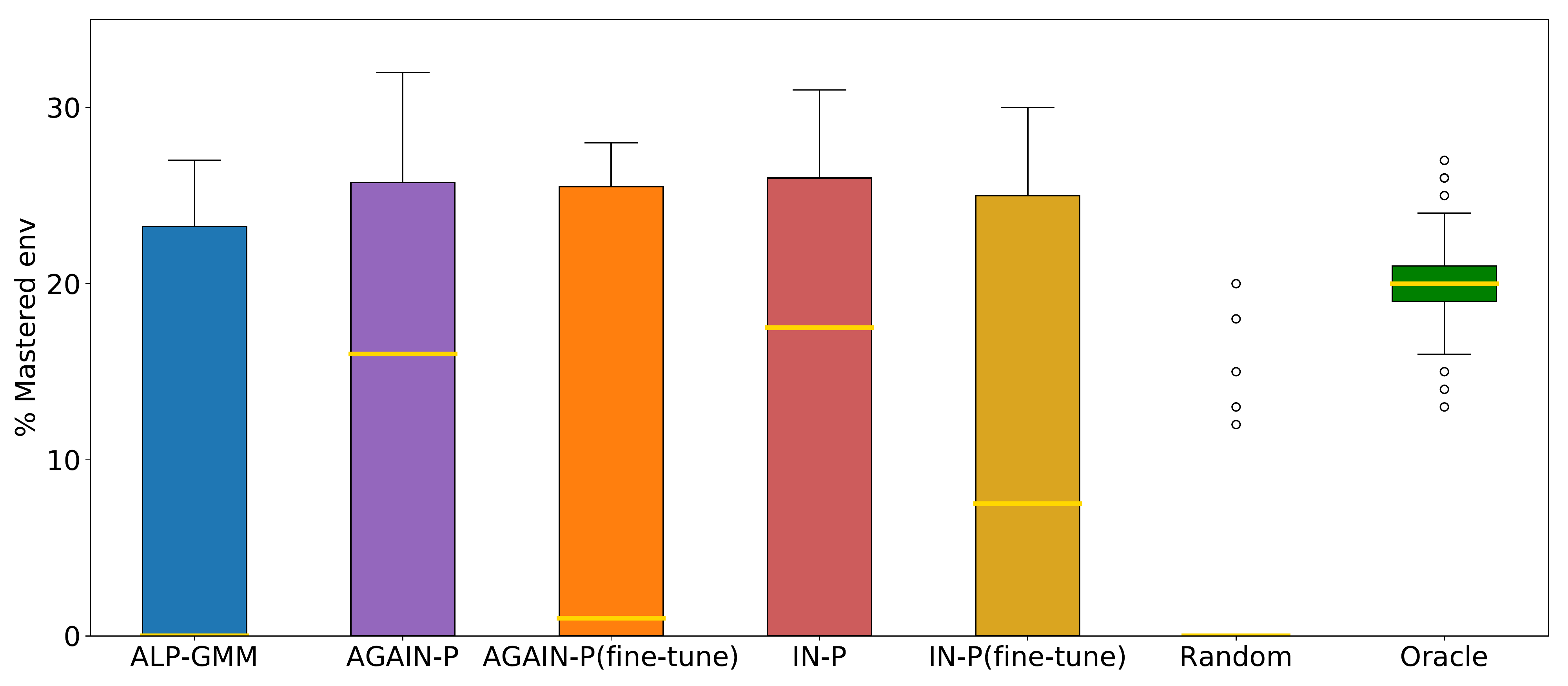}}

\subfloat[with \textbf{Time-based} IN]{\includegraphics[width=0.9\textwidth]{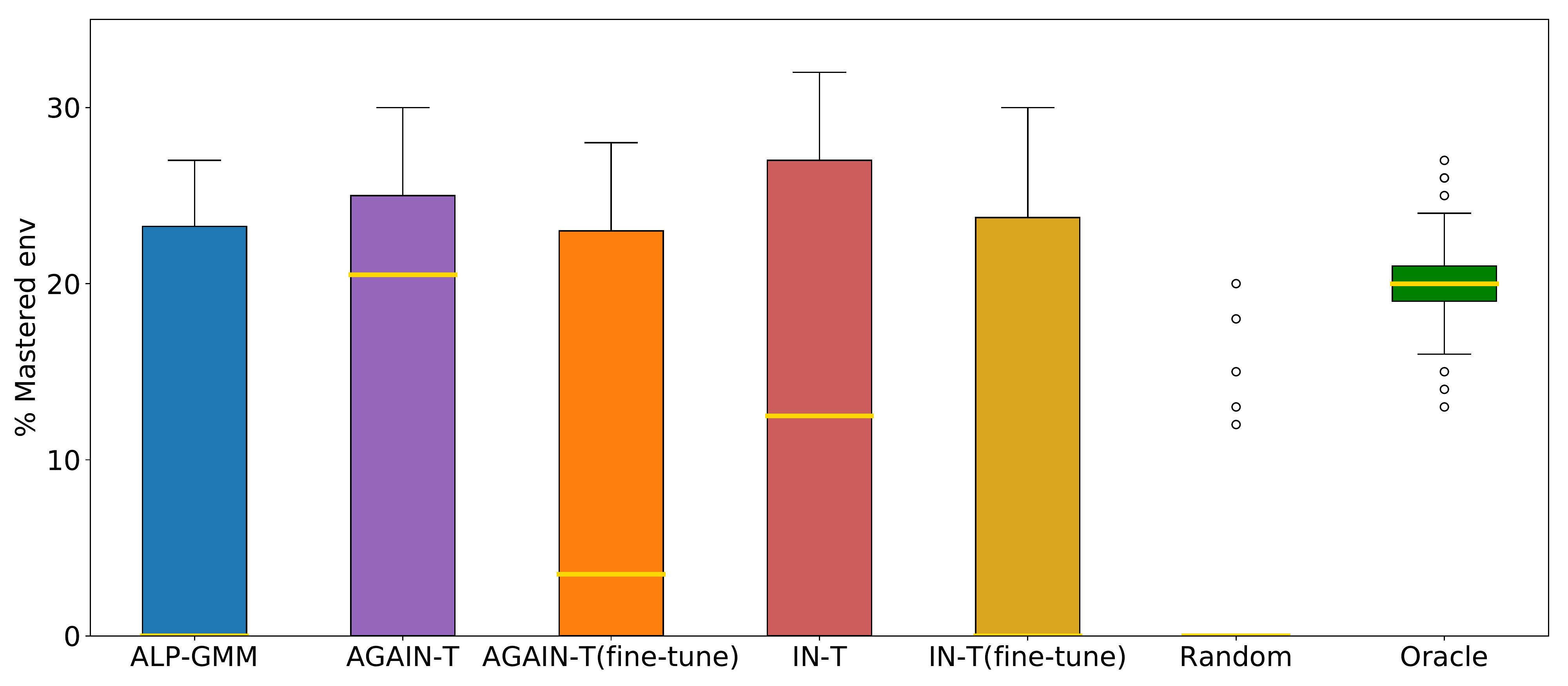}}

\subfloat[with \textbf{Reward-based} IN]{\includegraphics[width=0.9\textwidth]{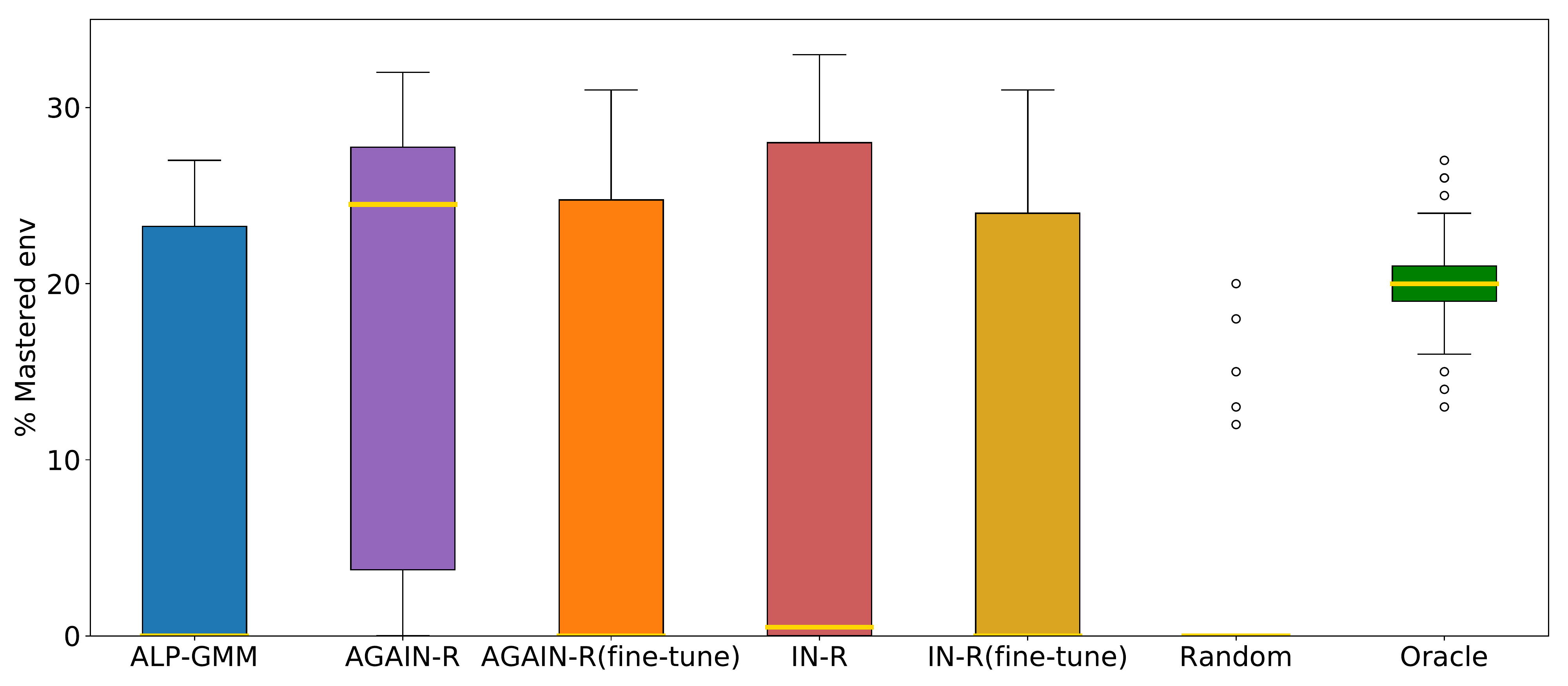}}
\caption{\footnotesize{\textbf{Box plots of the final performance of each condition with short bipedal walker after 20M environment steps.} Gold lines are medians, surrounded by a box showing the first and third quartile, which are then followed by whiskers extending to the last datapoint or $1.5$ times the inter-quartile range. Beyond the whiskers are outlier datapoints.} From top to bottom, each box plot presents results when using either Pool-based, Reward-based, or Time-based IN, respectively. }
\label{short-exps-boxplots}
\end{figure*}

\begin{figure*}[htb!]
\centering
\subfloat[\textbf{Pool-based} IN]{\includegraphics[width=0.33\textwidth]{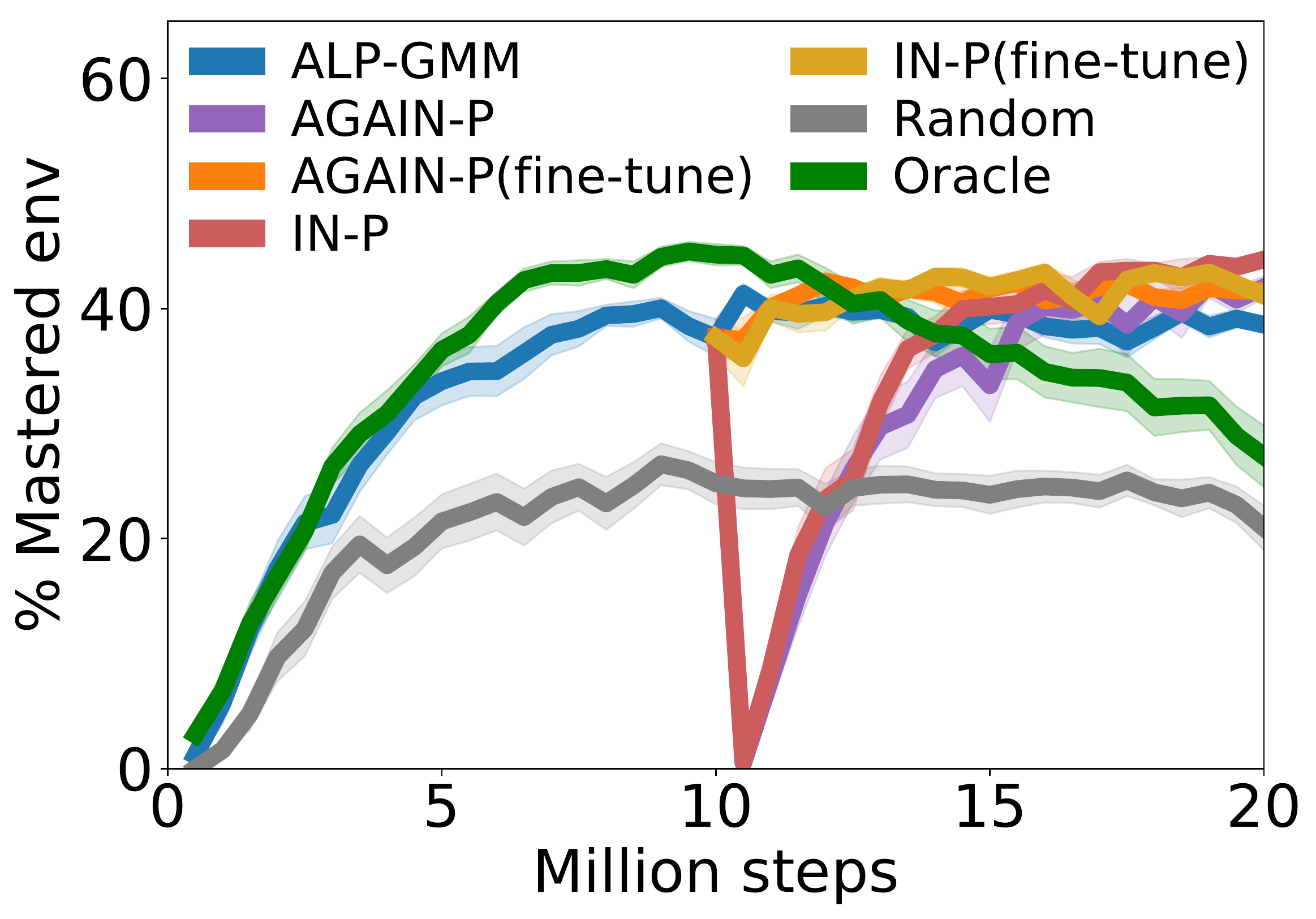}}
\subfloat[\textbf{Time-based} IN]{\includegraphics[width=0.33\textwidth]{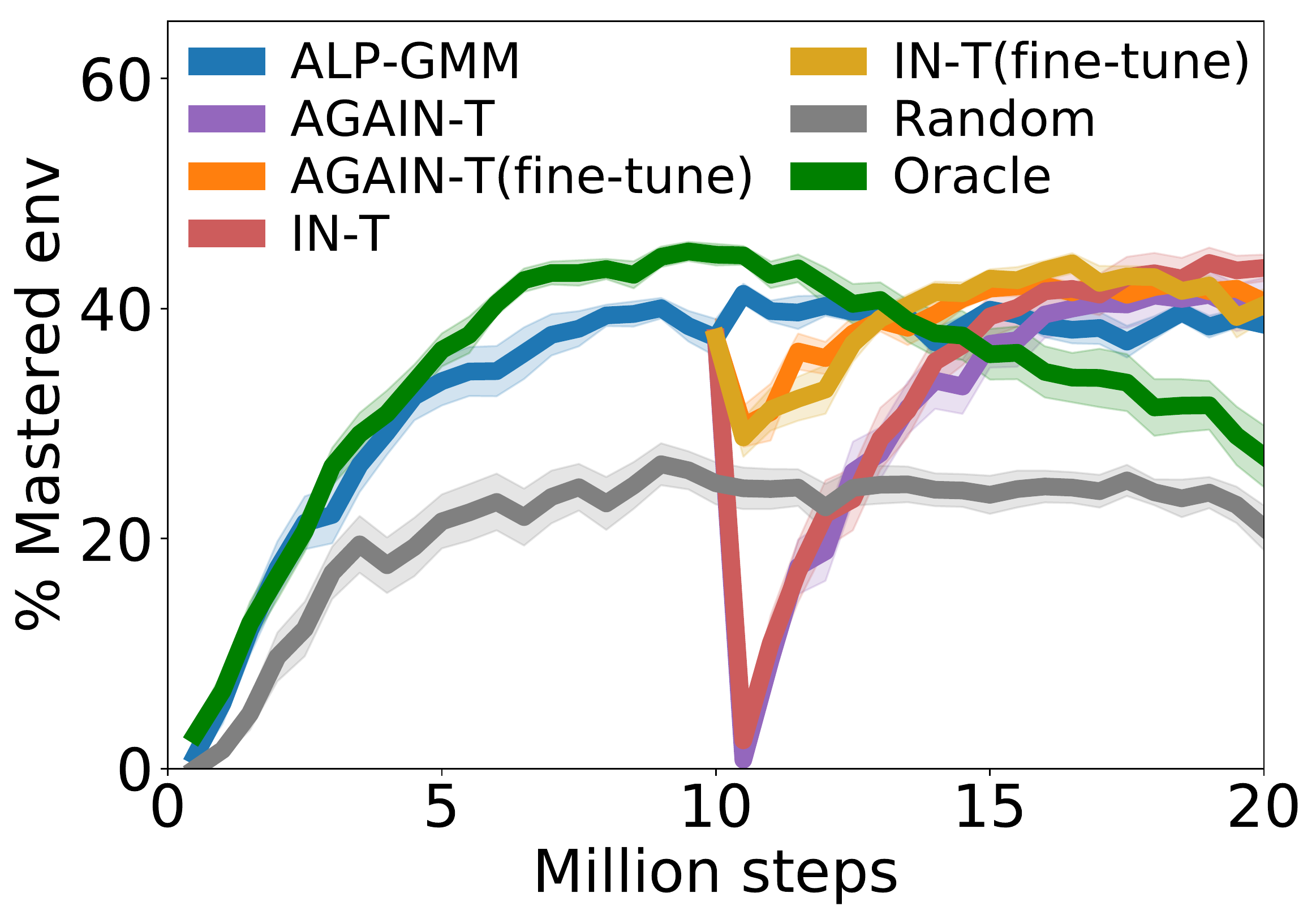}}
\subfloat[\textbf{Reward-based} IN]{\includegraphics[width=0.33\textwidth]{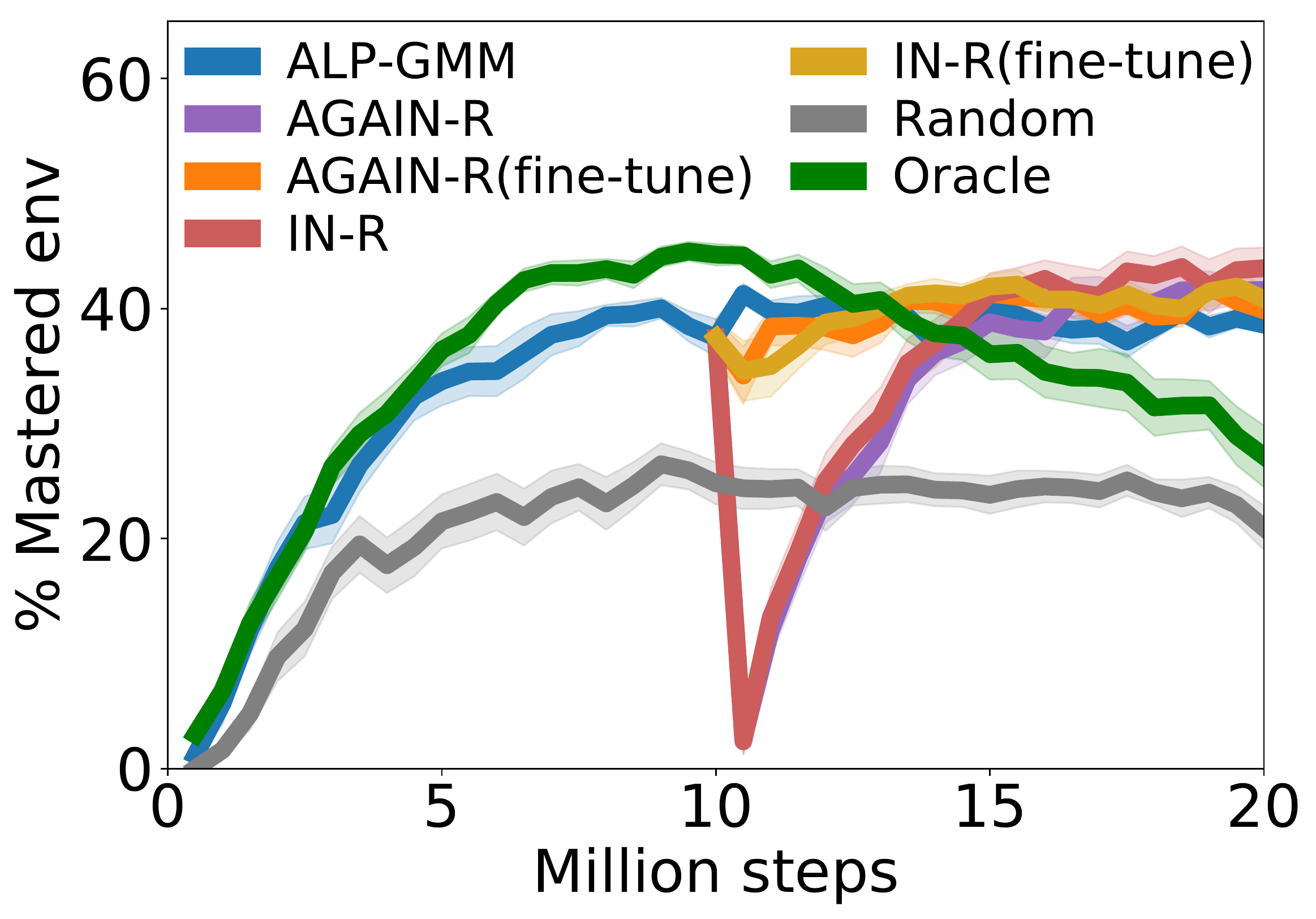}}
\caption{\footnotesize{\textbf{Evolution of performance across 20M environment steps of each condition with default bipedal walker.} Each point in each curve corresponds to the mean performance (30 seeds), defined as the percentage of mastered tracks (ie. $r>230$) on a fixed test set. Shaded areas represent the standard error of the mean.}}
\label{default-exps-curves}
\end{figure*}

\begin{figure*}[htb!]
\centering
\subfloat[\textbf{Pool-based} IN]{\includegraphics[width=0.33\textwidth]{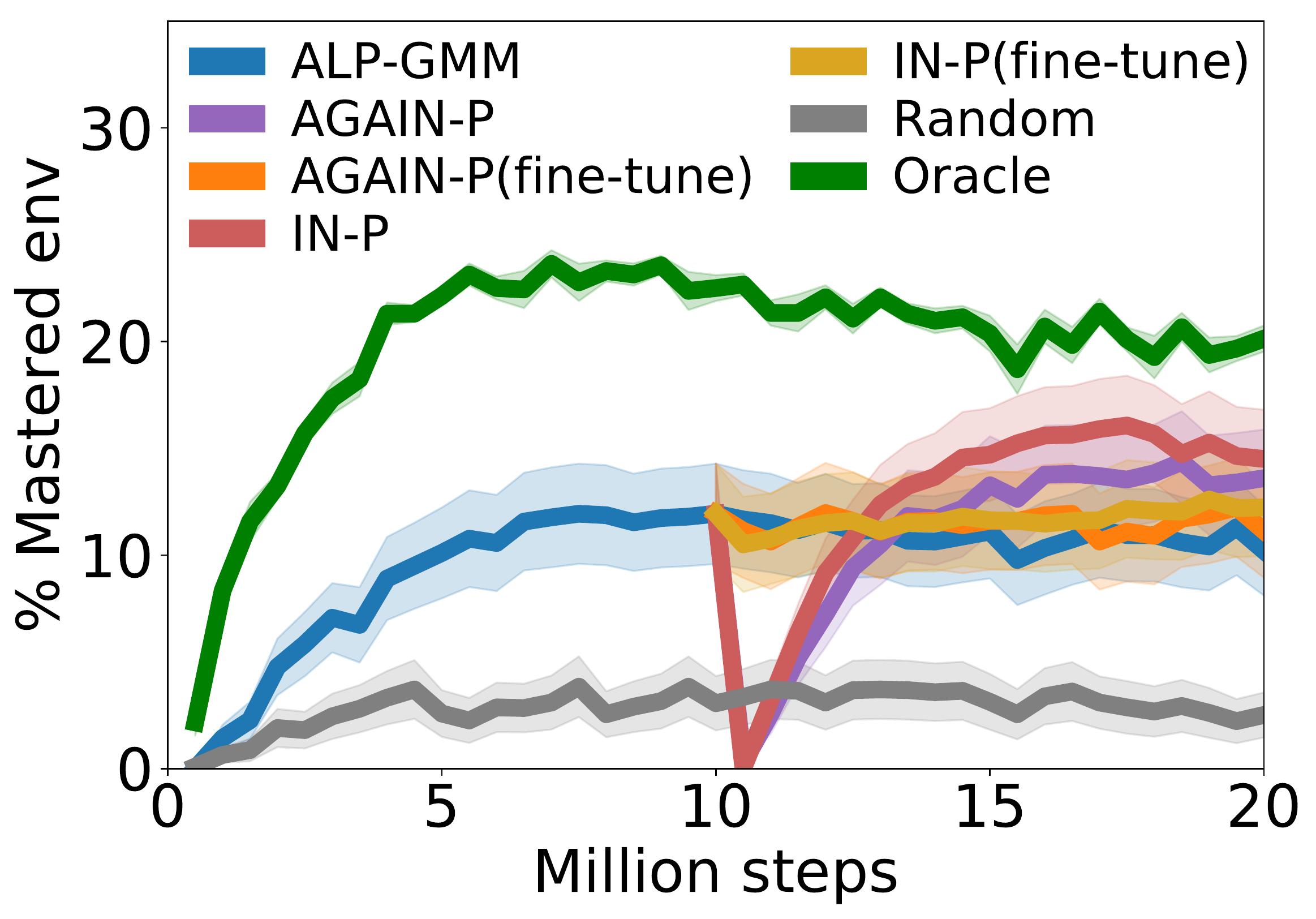}}
\subfloat[\textbf{Time-based} IN]{\includegraphics[width=0.33\textwidth]{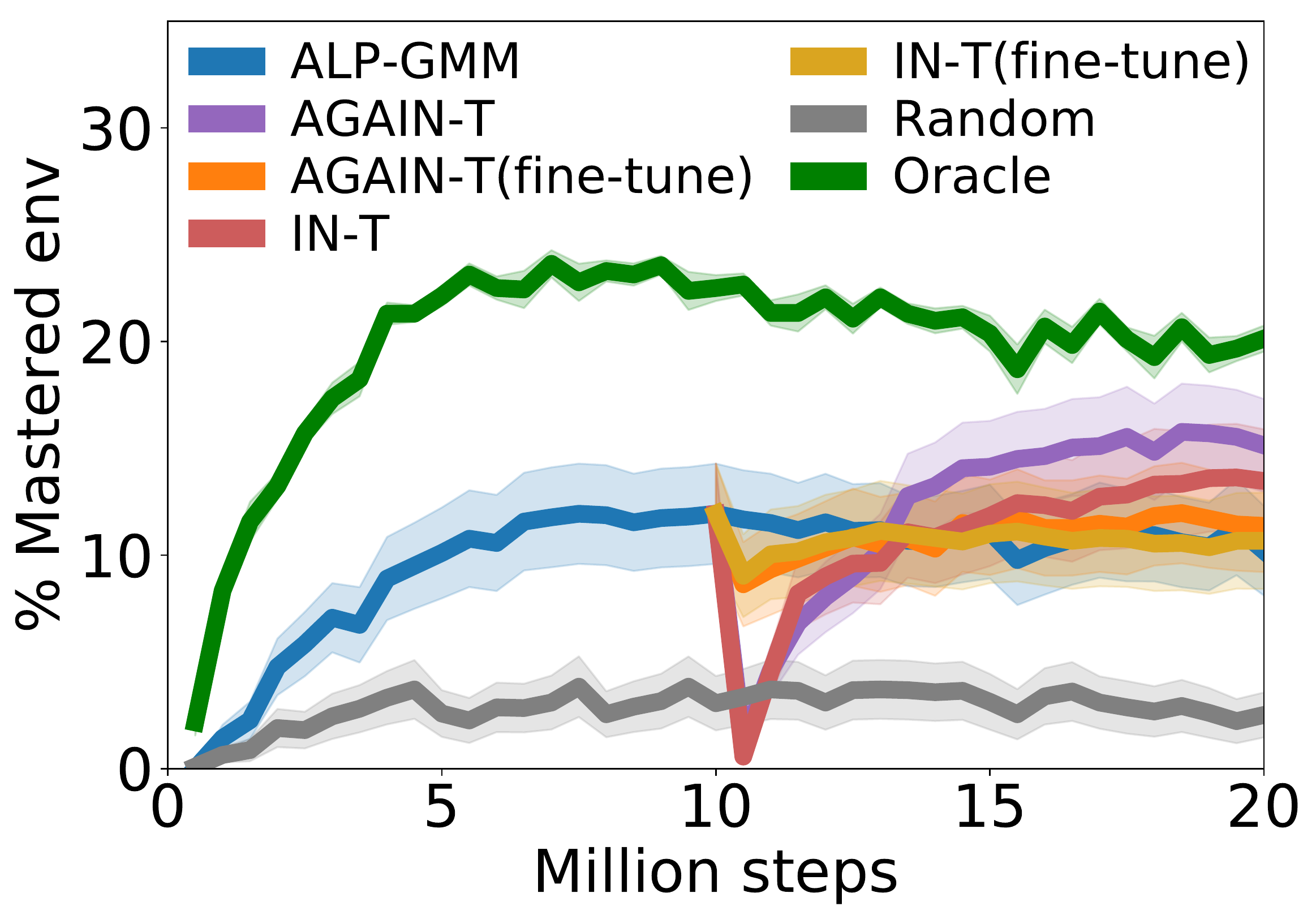}}
\subfloat[\textbf{Reward-based} IN]{\includegraphics[width=0.33\textwidth]{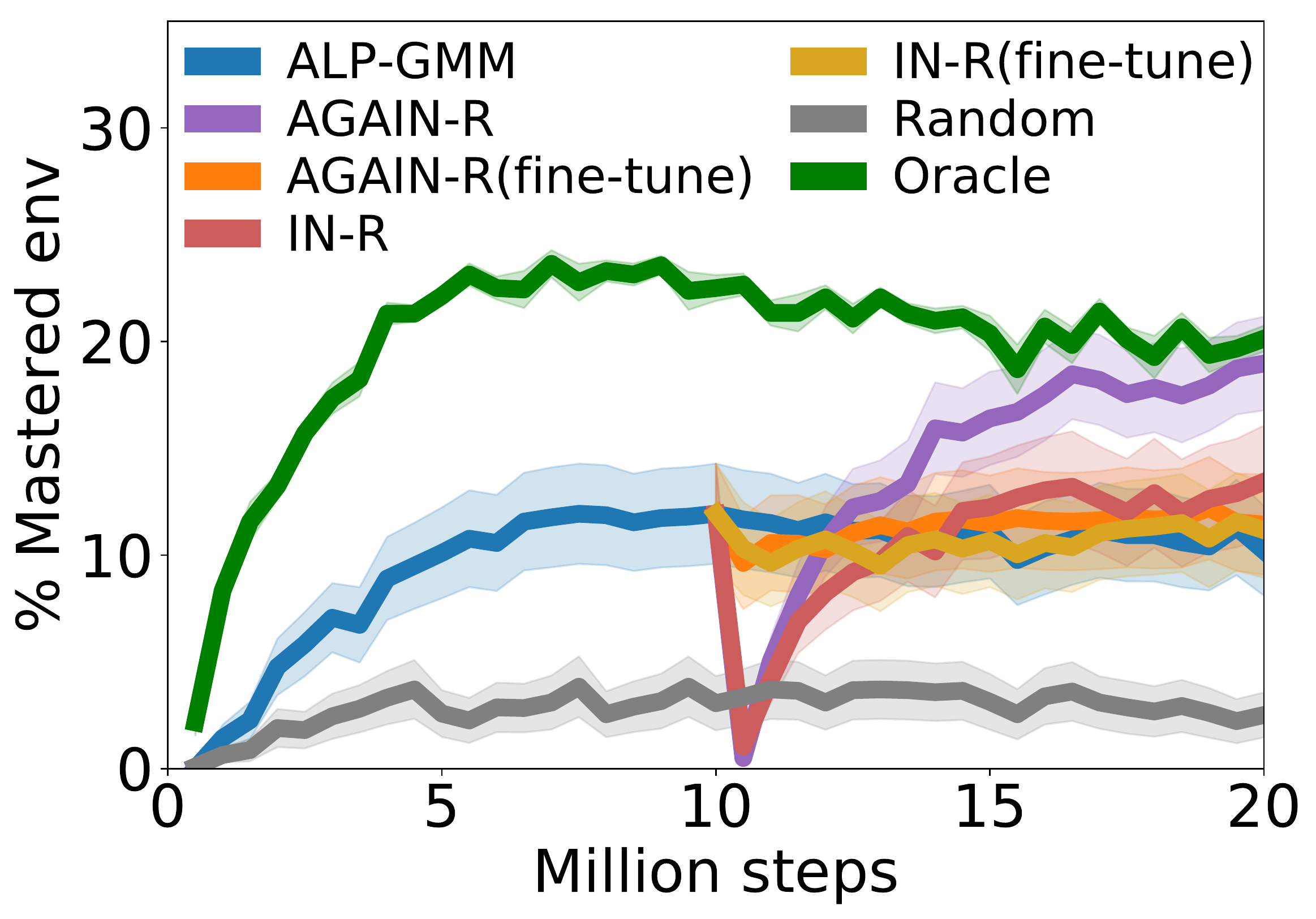}}
\caption{\footnotesize{\textbf{Evolution of performance across 20M environment steps of each condition with short bipedal walker.} Each point in each curve corresponds to the mean performance (30 seeds), defined as the percentage of mastered tracks (ie. $r>230$) on a fixed test set. Shaded areas represent the standard error of the mean.}}
\label{short-exps-curves}
\end{figure*}

\end{document}

%% file: math_commands.tex
\usepackage{amsmath,amsfonts,bm}

\def\eqref#1{equation~\ref{#1}}

\def\1{\bm{1}}

\DeclareMathAlphabet{\mathsfit}{\encodingdefault}{\sfdefault}{m}{sl}
\SetMathAlphabet{\mathsfit}{bold}{\encodingdefault}{\sfdefault}{bx}{n}

